# SliceBridge: Context-Consistent Repair of Corrupted Slice Intervals in T1-Weighted MRI

Jiheng Li*[a], Michael E. Kim[a], Trent M. Schwartz[b], Gaurav Rudravaram[b], Derek B. Archer[c], Timothy J. Hohman[c], for the Alzheimer's Disease Neuroimaging Initiative[1], Lianrui Zuo[b], Bennett A. Landman[a,b]

[a]Department of Computer Science, Vanderbilt University, Nashville, TN, USA; [b]Department of Electrical and Computer Engineering, Vanderbilt University, Nashville, TN, USA; [c]Vanderbilt Memory and Alzheimer's Center, Vanderbilt University Medical Center, Nashville, TN, USA; *Corresponding author: jiheng.li.1@vanderbilt.edu

## ABSTRACT

Structural magnetic resonance imaging (MRI) images are sometimes corrupted over a contiguous set of slices, where acquisition, motion, hardware, or reconstruction effects leave a single slice or short interval inconsistent with its neighbors while the rest of the image remains usable. Such localized corruption can bias downstream morphometric analysis, yet discarding or reacquiring an otherwise usable image is costly. We formulate this as an image restoration problem: given the location of the affected interval, reconstruct those slices from the surrounding anatomical and imaging context. We propose SliceBridge, a framework for restoring corrupted slice intervals in T1-weighted MRI using rectified flow matching conditioned on the surrounding intact slices and their relative slice positions. Through-plane consistency is encouraged by coupling the slices within the interval through interval-correlated initial noise, a shared flow time, and synchronized sampling. The restored interval is then inserted back, leaving all other slices unchanged. We trained and validated the model on 9,877 T1-weighted brain MRI volumes from four datasets and evaluated it on 581 external subjects using clean interval withholding and controlled corruptions. Compared with a matched model that reconstructed target slices independently, SliceBridge reduced error in slice-to-slice changes within repaired intervals by 32.9%-41.3% across interval lengths and achieved higher SSIM at every interval length. In controlled-corruption cases, SliceBridge reduced the median error in regional brain volume estimates produced by a downstream segmentation model from 1.95% in corrupted volumes to 1.05%.

## 1. INTRODUCTION

T1-weighted structural magnetic resonance imaging (MRI) is widely used in morphometry, brain segmentation, and other neuroimaging analyses, all of which depend on the reliable depiction of anatomical structure and regional anatomical boundaries. Acquisition and reconstruction abnormalities, however, alter those boundaries and bias estimates of brain volume or cortical anatomy[1–3]. Such abnormalities are not always distributed throughout the entire volume: a single axial slice or a short contiguous interval may appear inconsistent with its neighbors through abrupt changes in intensity, sharpness, noise, alignment, or anatomical appearance[4–6]. We refer to this as localized slice corruption (Figure 1). This corruption often results from acquisition abnormalities, motion, hardware, or reconstruction effects, although the reconstructed image alone may not reveal the underlying cause[5,7]. Although a corrupted interval typically occupies only a small portion of the volume, it may still disrupt downstream processing. Figure 1 shows a real example in which a corrupted interval coincides with localized fragmentation in a downstream whole-brain segmentation.

---

[1] Data used in preparation of this article were obtained from the Alzheimer's Disease Neuroimaging Initiative (ADNI) database (adni.loni.usc.edu). As such, the investigators within the ADNI contributed to the design and implementation of ADNI and/or provided data but did not participate in analysis or writing of this report. A complete listing of ADNI investigators can be found at: http://adni.loni.usc.edu/wp-content/uploads/how_to_apply/ADNI_Acknowledgement_List.pdf

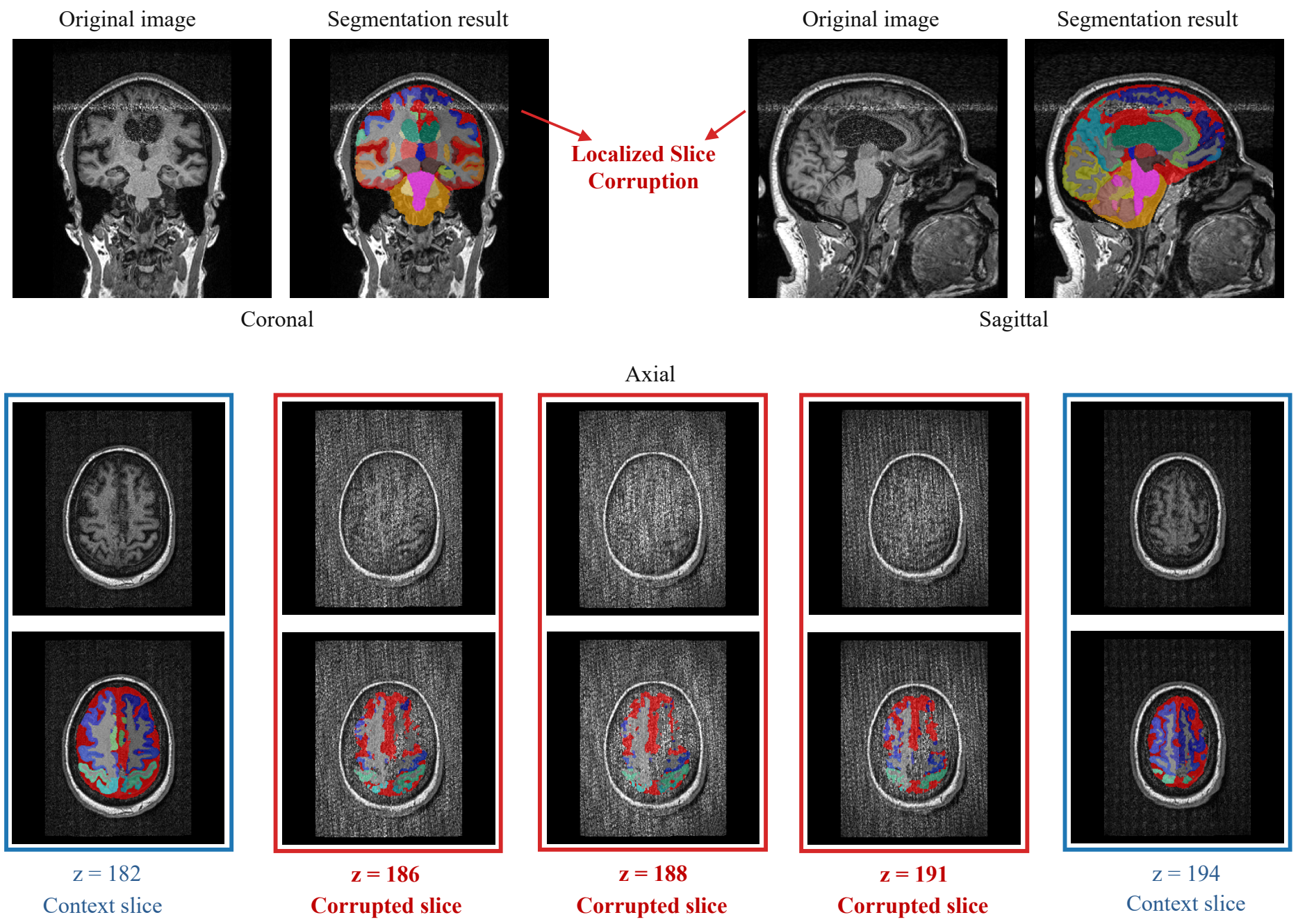

**Figure 1.** Real localized slice corruption spanning several adjacent slices in a T1-weighted volume. The corrupted slices (red) show local signal abnormalities and coincide with fragmentation in a downstream segmentation method (SLANT-TICV). Neighboring unaffected slices are shown in blue.

Localized slice corruption often comes with a trade-off: the affected interval cannot be trusted, but rejecting the entire scan would discard unaffected regions, while whole-volume correction would unnecessarily alter regions that remain usable[4,8,9]. Selective restoration confined to the corrupted region avoids both undesired outcomes. We therefore consider the setting in which a single axial slice or a contiguous axial interval of a T1-weighted volume has been identified as corrupted, either by manual quality control or by an automated slice-level detector. Because the corrupted content cannot be trusted, it is excluded from the conditioning input: the repair (restoration) is inferred only from the intact neighboring slices within the same volume, and all intact slices are left unchanged. Since the restoration replaces complete axial slices, it must satisfy two requirements. Each restored slice must be anatomically plausible in isolation, which we term in-plane (axial) reconstruction fidelity, and the restored slices together must form a coherent slice-to-slice progression within the interval and across its boundaries, which we term through-plane consistency.

Related work provides complementary precedents but not a direct solution. Prediction-based outlier replacement is established in diffusion MRI, where an unreliable slice can be replaced using an expectation inferred from other diffusion measurements[10,11]. This approach relies on redundancy across diffusion measurements, directions, and shells, which is unavailable within a single T1-weighted volume. Structural MRI studies instead show that same-volume spatial context can support localized reconstruction[6,12–14]. Verschuur et al. interpolated identified motion-affected slices in neonatal T2-weighted MRI[6], while Chai et al. restored missing through-plane content by inpainting masked rows in sagittal or coronal reformats, providing an indirect orthogonal-plane reconstruction formulation[13]. However, neither route is sufficient here. Interpolation satisfies through-plane consistency by construction but cannot synthesize anatomy that is not a blend of its neighbors, and the error grows with interval length; orthogonal-plane inpainting encourages smooth changes within each sagittal or coronal slice, but it recovers axial content only implicitly and does not directly model each complete axial slice or how the repaired slices change from one to the next. Together, these gaps motivate three questions. The first two questions follow from the repair requirements above: how direct reconstruction of complete axial slices compares with linear interpolation and with orthogonal-reformat inpainting, and whether coordinating the reconstruction of multiple slices improves through-plane consistency over reconstructing each slice independently. The third concerns downstream value: whether selective restoration reduces corruption-induced error in morphometric measurements.

We introduce SliceBridge to address these questions through direct axial reconstruction with interval-level coordination. Given a corrupted interval, SliceBridge reconstructs each complete axial slice using a shared 2D U-Net[15] trained with rectified flow matching[16], conditioned on the intact neighboring slices and on their positional information. In-plane plausibility alone, however, does not deliver through-plane consistency: independent stochastic trajectories can introduce abrupt changes between adjacent predictions[17,18]. SliceBridge therefore couples generation using interval-correlated initial noise, an interval-shared flow time during training, and synchronized sampling during inference. To isolate what this coupling contributes, we compare against an uncoupled-flow model matched in data, context, positional conditioning, and architecture but generating each slice independently. We then evaluate whether the resulting restoration reduces corruption-induced error in downstream morphometric measurements.

## 2. METHOD

SliceBridge restores a corrupted interval by reconstructing each of its axial slices directly from the surrounding intact anatomy, and by coupling those reconstructions so that they form a coherent through-plane progression. Figure 2 summarizes the SliceBridge framework.

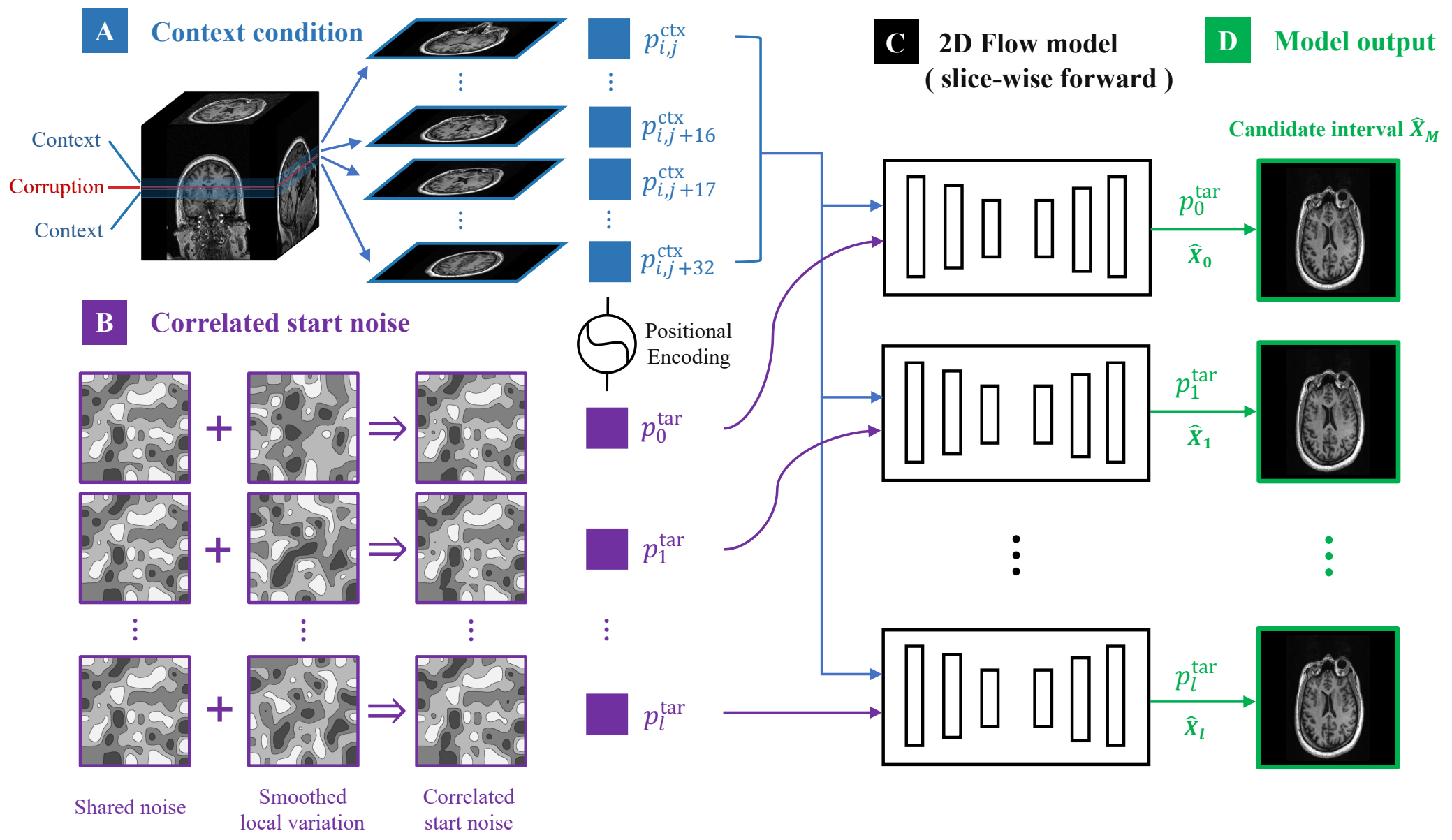

**Figure 2.** SliceBridge framework. (A) Intact context slices are paired with the context-offset map. (B) For each target slice, a shared Gaussian field and smoothed slice-specific variation form interval-correlated initial noise, which is paired with its target-position map. (C) A shared 2D rectified-flow model reconstructs the targets using synchronized interval timing. (D) The model generates all target slices conditioned on their corresponding target-position map.

### 2.1 Direct axial reconstruction

Consider a T1-weighted MRI volume in which a contiguous interval of $m$ axial slices has been identified as corrupted. Let $i \in \{0, \dots, \mathrm{m}-1\}$ index the slices within the interval and let $z_i$ denote the axial index of slice $i$ in the volume.

**Context.** SliceBridge conditions on the intact slices flanking the interval, which we refer to as the **context.** Because the reconstruction must reproduce both the local anatomy and the imaging characteristics of the volume, SliceBridge collects 16 context slices from each side of the interval, denoted $C = \{c_j\}$, where $c_j$ denotes the axial index of context slice $j$. If another corrupted interval falls within the search range, those slices are excluded and the search continues outward until 16 slices per side have been retained. The retained context is therefore not necessarily contiguous, and the axial distance between a context slice and a target slice is not fixed.

**Positional conditioning.** Every target slice in the interval is reconstructed from the same shared context stack, so the geometric relationship between the context and the current target slice differs from one target to the next, and the image channels alone do not encode this relationship. SliceBridge supplies the missing geometry with two positional maps: **context-offset maps**, which describe where each context slice lies relative to the current target, and a **target-position map**, which records where the target lies within the interval.

The context-offset map is defined as

$$p_{i,j}^{\mathrm{ctx}} = \frac{c_j - z_i}{16}. \tag{1}$$

The denominator 16 is a fixed normalization scale. The sign indicates which side of the target slice contains the context slice, while the magnitude records its normalized distance.

For target $i$, the target-position map is defined as

$$p_i^{\mathrm{tar}} = \begin{cases} 0, & m = 1, \\ 2\dfrac{i}{m-1} - 1, & m > 1. \end{cases} \tag{2}$$

The value ranges from $-1$ at the first target slice to $+1$ at the last, with values near zero identifying targets closer to the interval center, which are farthest from observed context on either side. For $m = 1$, the value is zero.

**Reconstruction model.** Rectified flow[16] is used to generate the axial target slice. Let $y_i$ denote the clean target image and $\epsilon_i$ its initial noise. At flow time $t \sim \mathcal{U}(0,1)$, the intermediate state is

$$x_i(t) = (1-t)\epsilon_i + t y_i, \tag{3}$$

and the corresponding flow velocity is

$$u_i = y_i - \epsilon_i. \tag{4}$$

The velocity network $v_\theta$ receives the current state $x_i(t)$, the flow time $t$, the context images $C$, the context-offset maps $p_{i,j}^{\mathrm{ctx}}$ for all context slices $j$, and the target-position map $p_i^{\mathrm{tar}}$. For each target slice $i$, it is trained using the flow-matching loss

$$\ell_i = \left\| v_\theta(x_i(t), t, C, p_{\mathrm{i,j}}^{\mathrm{ctx}}, p_{\mathrm{i}}^{\mathrm{tar}}) - u_i \right\|_2^2. \tag{5}$$

The same 2D U-Net[15] parameters are shared across all target slices.

This formulation reconstructs each target slice as a complete axial image, but it does not ensure that the slice-wise predictions form a coherent interval. Adjacent target slices may begin from unrelated noise and follow unrelated generative trajectories, so the reconstructed slices can be individually plausible while showing abrupt changes in intensity or fine anatomical structure when stacked[18].

### 2.2 Interval coordination

SliceBridge couples target generation across the corrupted interval at three points: the initial noise, the flow time used during training, and the sampling process used during inference.

**Interval-correlated noise initialization.** Independent initial noise gives adjacent target slices unrelated starting states. Motivated by correlated noise priors for ordered video frames[17], SliceBridge instead constructs the initial noise jointly across the interval. Let $g$ denote a single Gaussian field shared by all target slices and let $e = (e_0, \dots, e_{m-1})$ denote an ordered set of independent Gaussian fields for each target slice. After the fields in $e$ are smoothed along the interval using a three-slice moving average $S_3$, the initial noise for target slice $i$ is

$$\epsilon_i = \frac{w_s g + w_l [S_3(e)]_i}{\sqrt{w_s^2 + w_l^2}}. \tag{6}$$

Here, $w_s$ and $w_l$ weight the shared and slice-specific components, respectively; we set $w_s = 1.0$ and $w_l = 0.3$. The shared component ties the starting states together across the interval; the smoothed slice-specific component keeps individual targets distinct while allowing the noise to vary gradually from one slice to the next.

**Interval-shared flow time.** Correlated starting states alone do not ensure that the target slices are trained at the same stage of generation. Under independently sampled flow times, different target slices from the same interval would be supervised at different points along their respective trajectories from noise to clean image. SliceBridge therefore samples at a single flow time $t$ per interval during training and applies it to every target. The $t$ used in Equations (3) and (5) is thus shared across all targets in the interval.

**Synchronized sampling.** The same alignment must also be preserved when the repair candidate is generated. During inference, all target slices use the same Euler time grid and are updated together at every sampling step. No target slice therefore advances ahead of or behind the others during generation.

### 2.3 Center-weighted flow objective

The three mechanisms above couple the generative trajectories. A separate training consideration is that reconstruction difficulty varies with target slice position. Target slices near the interval boundaries are close to observed context, whereas target slices near the center are farther from context on both sides. SliceBridge therefore uses a center-weighted flow objective for intervals of six or more slices. Using the target-position value $p_i^{\text{tar}}$ from Equation (2), the weights are defined as

$$w_i = \frac{1 + \alpha(1 - |p_i^{\text{tar}}|)}{\frac{1}{m}\sum_{k=0}^{m-1}[1 + \alpha(1 - |p_k^{\text{tar}}|)]}, \tag{7}$$

where $\alpha = 1$. The weights increase toward the interval center and are normalized to unit mean, so supervision is redistributed across target positions without changing the overall loss scale. Using the slice-wise loss $\ell_i$ from Equation (5), the final interval flow matching loss is

$$\mathcal{L}_{\text{flow}} = \frac{1}{m}\sum_{i=0}^{m-1} w_i\, \ell_i. \tag{8}$$

For shorter intervals, all $w_i$ are set to one, recovering the ordinary mean flow-matching loss.

### 2.4 Training and inference

During training, contiguous intervals of lengths $\{1,2,4,6,8,10,12\}$ are withheld from clean volumes. The withheld slices are used only as target images, while the model receives the context outside the withheld interval. To expose the model to context gaps like those created when other corrupted slices are excluded, eligible context candidates are additionally skipped with probability 0.1 during training. Training intervals were sampled only when 16 eligible context slices could be obtained on both sides.

At inference, the generated target slices are assembled in axial order to form the repair candidate. The candidate is inserted only at the corrupted locations.

## 3. EXPERIMENTAL SETUP

### 3.1 Data and preprocessing

We used T1-weighted adult brain MRI from ADNI[19], AOMIC[20–23], NACC[24], and OASIS-3[25] for model development and reserved IXI[26] exclusively for external evaluation. From each subject in the development datasets, we selected one scan from a randomly chosen session, giving 10,131 volumes. All volumes underwent identity-duplication auditing using HAPPEN[27] and manual visual quality assessment using the workflow of Kim et al.[28], and 9,877 were retained. For model development, we split the development cohort by subject, yielding 7,901 training and 1,976 validation volumes. For IXI, we likewise selected one T1-weighted volume per subject, yielding 581 volumes, all of which passed quality assessment. Table 1 summarizes the cohorts.

Table 1. Development and external evaluation cohorts after quality control.

| Dataset | Training | Validation | External evaluation |
|---|---|---|---|
| ADNI | 2,479 | 619 | -- |
| AOMIC | 1,098 | 271 | -- |
| NACC | 3,609 | 911 | -- |
| OASIS-3 | 715 | 175 | -- |
| IXI | -- | -- | 581 |
| **Total** | **7,901** | **1,976** | **581** |

Data used in the preparation of this article were partially obtained from the Alzheimer's Disease Neuroimaging Initiative (ADNI) database (adni.loni.usc.edu). The ADNI was launched in 2003 as a public-private partnership, led by Principal Investigator Michael W. Weiner, MD. The primary goal of ADNI has been to test whether serial magnetic resonance imaging (MRI), positron emission tomography (PET), other biological markers, and clinical and neuropsychological assessment can be combined to measure the progression of mild cognitive impairment (MCI) and early Alzheimer's disease (AD).

**Preprocessing.** All retained volumes underwent the same preprocessing. We reoriented each volume to the closest canonical orientation, clipped nonzero foreground intensities at the 0.5th and 99.5th percentiles, and scaled them to [0, 1] using min-max normalization. Each volume was then represented as an axial slice sequence, and each slice was center-cropped or zero-padded to 256 × 256 pixels. No image interpolation or through-plane resampling was used.

### 3.2 Evaluation design and protocols

The evaluation was designed to address the three research questions introduced above. Reconstruction fidelity assessed how closely the reconstructed target slices matched their clean references and was used to compare the alternative reconstruction formulations. Through-plane consistency assessed whether the reconstructed slices formed a coherent slice-to-slice progression within the corrupted interval and across its boundaries, thereby testing the interval-coordination design. Downstream morphometry assessed whether repair reduced corruption-induced error in regional volume measurements. The first two evaluations required known clean target slices and were therefore performed by withholding clean intervals from the external IXI volumes. Downstream morphometry required paired clean, corrupted, and repaired volumes and was therefore evaluated using controlled corruptions, with the original IXI volumes retained as references.

**Clean-interval withholding.** We constructed intervals of lengths {1, 2, 4, 6, 8, 10, 12} using the same bilateral-context requirements as during training. The withheld slices were retained as clean references but were unavailable to the repair methods. Only intervals for which 16 eligible intact slices could be obtained on both sides were included; intervals near either end of the axial sequence were therefore excluded when the full context could not be formed.

**Controlled-corruption morphometry.** Because real-world localized slice corruptions lack verified uncorrupted counterparts, their effects on morphometric measurements cannot be quantified directly. We therefore applied one of seven controlled corruption types to copies of the clean IXI volumes while retaining the original preprocessed volumes as references: signal dropout, pile-up-like intensity elevation, whiteout, Rician-like noise, Gaussian blur, in-plane rigid motion, or RF-like banding. The corruptions were applied after preprocessing so that each corrupted volume remained spatially aligned with its clean reference and used the same intensity scale. Corruptions were placed in one to three nonoverlapping axial intervals sampled from the central foreground-containing slice range, and their combined length was limited to 5% of the slices in that range. Allowed interval lengths were 1-2 for whiteout; 1-4 for pile-up and RF-like banding; 1-6 for motion; 1-8 for dropout and blur; and 1-12 for noise. Each interval was also required to have 16 eligible intact slices available on both sides.

**Common evaluation conditions.** Across both protocols, all repair methods were evaluated at the same corrupted interval locations and received no image content from the corrupted interval. In the controlled-corruption protocol, the learned models were not given the corruption type or the corrupted target content. If a volume contained multiple corrupted intervals, all corrupted slices were excluded from the context set and each contiguous interval was reconstructed separately.

### 3.3 Comparison methods

We compared four repair methods on the IXI evaluation cases: linear interpolation, EG-GAN, Uncoupled flow, and SliceBridge. Linear interpolation and EG-GAN represented the interpolation and orthogonal-plane inpainting alternatives in the first research question. The matched comparison between Uncoupled flow and SliceBridge isolated the effect of interval-level coordination in the second research question.

Linear interpolation estimated the corrupted slices from the nearest intact slice on each side of the interval, with weights set by axial position. This provided a simple deterministic baseline based only on the two interval boundaries.

EG-GAN adapted the edge-guided adversarial restoration method of Chai et al.[13] to the corrupted-interval task. Each T1-weighted volume was reformatted into coronal sections, in which the corrupted axial interval appeared as contiguous masked rows. EG-GAN completed the edge maps and image intensities within these rows; the restored sections were then mapped back to axial space, and only the corrupted slices were replaced. We used a single coronal orientation without multi-view fusion. EG-GAN was trained using the same development volumes and withheld target intervals as the flow models, with each axial interval represented as contiguous masked rows in the coronal reformats.

Uncoupled flow used the same training data, context images, positional encodings, rectified-flow formulation, and 2D U-Net as SliceBridge, but omitted all interval-level components described in Section 2.2. Each target slice used independent initial noise and an independently sampled flow time during training, and the model was optimized with the ordinary unweighted flow-matching loss.

SliceBridge used the complete model described in Section 2.

**Implementation details.** The validation split was used for hyperparameter and checkpoint selection for all learned models. The two flow models used the same time-conditioned 2D U-Net with a base width of 64 and channel multipliers of (1, 2, 4, 8). Both were trained for 50,000 AdamW steps with an initial learning rate of $10^{-4}$and used 10 Euler steps at inference. For each flow model, one repair candidate was generated per interval using a deterministic case-specific seed, with no repeated sampling or candidate selection.

### 3.4 Evaluation metrics

**Reconstruction fidelity.** We first measured how closely each reconstructed slice matched its clean reference using the foreground structural similarity index (SSIM) and foreground peak signal-to-noise ratio (PSNR). The evaluation mask contained pixels whose intensity in the clean reference slice exceeded $10^{-4}$, so every method was scored over the same foreground support while zero-valued padding was excluded. Because all images were normalized to [0, 1], the data range used for SSIM and PSNR was set to 1. Slice-level values were averaged within each corrupted interval, yielding one score per subject, method, and interval length. We reported these interval-level scores separately by interval length. To examine how reconstruction performance varied with distance from the observed boundaries, we also analyzed slice-level SSIM by relative position within intervals of length 12.

**Through-plane consistency.** Through-plane consistency was measured at two levels. Internal z-gradient MAE assessed slice-to-slice progression between adjacent reconstructed slices within the corrupted interval, whereas boundary z-gradient MAE assessed consistency at the two interfaces between the reconstructed interval and the neighboring intact slices. Internal z-gradient MAE was evaluated only for intervals containing at least two slices, and both metrics were reported separately by interval length.

For each evaluated transition, we compared the change between two adjacent slices in the repaired volume with the corresponding change in the clean volume. For a set of evaluated slice-to-slice transitions $T$, the z-gradient MAE was defined as

$$\mathrm{zMAE}(T) = \frac{1}{|T| \times 256 \times 256} \sum_{z \in T} \|(\hat{x}_z - \hat{x}_{z-1}) - (x_z - x_{z-1})\|_1, \tag{9}$$

where $x_z$and $\hat{x}_z$ denote the clean and repaired slices at axial position $z$, respectively, and $T$ contains either the transitions within the corrupted interval or the two transitions at its boundaries. The absolute difference was averaged over all pixels and then across the evaluated transitions.

**Downstream morphometry.** We applied SLANT-TICV[29], a deep-learning whole-brain segmentation pipeline for T1-weighted MRI, to the clean and corrupted volumes and to the outputs of linear interpolation, EG-GAN, Uncoupled flow, and SliceBridge, using the same settings. The segmentation of the clean volume served as the reference. An ROI was considered affected when its clean segmentation overlapped a corrupted interval by at least 10 voxels. We also required a clean ROI volume of at least 100 $mm^3$ to avoid unstable percentage errors for very small structures; background and aggregate labels were excluded. Because corrupted intervals were not required to intersect a SLANT-TICV ROI, 133 cases had no eligible affected ROI and were excluded, leaving 448 cases for morphometric analysis.

For each affected ROI, we computed the absolute percentage error (APE) between its volume in a corrupted or repaired image and the corresponding clean volume:

$$\mathrm{APE}_r = 100 \times \frac{|V_r - V_r^{\mathrm{ref}}|}{V_r^{\mathrm{ref}}}. \tag{10}$$

In the formula, $V_r$ denotes the ROI volume measured from the corrupted or repaired image, and $V_r^{\mathrm{ref}}$ denotes the corresponding clean volume. For each case, we summarized each condition by the median APE across affected ROIs.

**Statistical analysis.** For each fidelity and through-plane metric, we compared SliceBridge with linear interpolation, EG-GAN, and Uncoupled flow using two-sided Wilcoxon signed-rank tests on paired IXI cases. Tests were performed separately at each interval length for which the metric was defined. Holm correction was applied separately within each metric across all interval lengths and the three method comparisons and Holm-adjusted $p < 0.05$ was considered significant. For morphometry, the same test compared SliceBridge with the corrupted volumes and the outputs of linear interpolation, EG-GAN and Uncoupled flow; Holm correction was applied across these four comparisons. To complement the hypothesis tests, we report win rates for the two primary comparisons associated with interval coordination and downstream recovery: SliceBridge versus Uncoupled flow for through-plane consistency, and SliceBridge versus the corrupted volumes for downstream morphometry. For each comparison, the win rate was defined as the proportion of paired cases in which SliceBridge achieved the preferred metric value. All confidence intervals were estimated by case-level bootstrap resampling and are reported as 95% confidence intervals.

# 4. RESULTS

Table 2. Quantitative evaluation on the external IXI cohort. (A) Reconstruction fidelity and through-plane consistency at representative short, medium, and long intervals, reported as means across 581 paired cases. (B) Median (interquartile range) affected-ROI APE across 448 controlled-corruption cases with at least one eligible affected ROI. The final column in each panel reports Holm-adjusted p-values from paired Wilcoxon signed-rank tests against SliceBridge. Bold indicates the best value.

**A. Clean-interval withholding**

| Metric | *m* | Linear Interpolation | EG-GAN | Uncoupled flow | SliceBridge | Holm-adjusted *p* |
|---|---|---|---|---|---|---|
| | 1 | 0.931 | 0.931 | 0.955 | **0.960** | |
| Foreground SSIM ↑ | 6 | 0.706 | 0.755 | 0.814 | **0.824** | $p < 0.001$ vs all |
| | 12 | 0.584 | 0.638 | 0.719 | **0.733** | |
| | 1 | 28.43 | 28.83 | 32.16 | **32.53** | |
| Foreground PSNR (dB) ↑ | 6 | 20.83 | 21.86 | 23.92 | **24.00** | $p < 0.001$ vs all |
| | 12 | 18.62 | 19.28 | 21.61 | **21.64** | |
| | 2 | 0.00976 | 0.01168 | 0.01241 | **0.00757** | |
| Internal z-gradient MAE ↓ | 6 | 0.01639 | 0.01915 | 0.02159 | **0.01404** | $p < 0.001$ vs all |
| | 12 | 0.01748 | 0.02256 | 0.02673 | **0.01570** | |
| | 1 | 0.01017 | 0.01050 | 0.00826 | **0.00744** | |
| Boundary z-gradient MAE ↓ | 6 | 0.01745 | 0.01666 | 0.01396 | **0.01348** | $p < 0.001$ vs all |
| | 12 | 0.01778 | 0.01775 | 0.01482 | **0.01389** | |

**B. Controlled-corruption morphometry**

| Condition | Median affected-ROI APE (%) | Holm-adjusted *p* vs SliceBridge |
|---|---|---|
| Corrupted | 1.95 (1.14-4.18) | $p < 0.001$ |
| Linear interpolation | 1.30 (0.87-2.17) | $p < 0.001$ |
| EG-GAN | 1.19 (0.81-1.78) | $p < 0.001$ |
| Uncoupled flow | **1.05 (0.71-1.45)** | $p = 0.370$ |
| SliceBridge | **1.05 (0.76-1.45)** | Reference |

## 4.1 Reconstruction fidelity

Direct axial reconstruction provided higher target fidelity than linear interpolation and coronal EG-GAN: both flow models achieved higher mean foreground SSIM and PSNR at every evaluated interval length (Table 2; Figure 3A). SliceBridge also exceeded each comparator for both metrics at every length, and all paired comparisons remained significant after Holm correction (all adjusted $p < 0.001$).

To further examine how reconstruction performance varied within a long interval, we analyzed the SSIM of each target slice by its relative position in intervals of length 12 (Figure 3B). Across methods, SSIM was highest near the observed boundaries and lowest near the interval center. The difference between SliceBridge and Uncoupled flow was also largest at the central positions, where the targets were farthest from observed context.

## 4.2 Through-plane consistency

Interval coordination improved both internal and boundary through-plane consistency relative to matched independent generation. Across intervals of length 2-12, SliceBridge reduced mean internal z-gradient MAE by 32.9%-41.3% relative to Uncoupled flow and produced lower error in all 581 paired cases at every length (Figure 3C). Because internal z-gradient MAE measures deviation from the corresponding clean slice-to-slice transition rather than from zero, it does not inherently favor smoother outputs: both attenuated and exaggerated transitions increase error. Consistent with this, linear interpolation, which enforces smooth through-plane variation, still had higher internal z-gradient MAE than SliceBridge at every interval length. Every comparison remained significant after Holm correction (all adjusted $p < 0.001$).

EG-GAN had lower mean internal error than Uncoupled flow, but both remained above linear interpolation, whereas SliceBridge was the only learned method to outperform interpolation across all evaluated lengths. High target fidelity alone therefore did not ensure a coherent repaired interval. At the two interval boundaries, both flow models achieved lower mean error than interpolation and EG-GAN. SliceBridge further reduced boundary error by 3.4%-10.0% relative to Uncoupled flow, with significant paired differences at every length (all adjusted $p < 0.001$).

### 4.3 Downstream morphometry

SliceBridge reduced corruption-induced morphometric error in the affected ROIs. Compared with the corrupted volumes, SliceBridge produced a median paired reduction of 0.81 percentage points in affected-ROI APE (95% CI, 0.64-0.96) and yielded lower error in 79.7% of paired cases (Figure 3D; adjusted $p < 0.001$). Linear interpolation and EG-GAN provided intermediate recovery, and SliceBridge significantly outperformed both methods after Holm correction (both adjusted $p < 0.001$).

No significant difference was detected between SliceBridge and Uncoupled flow (adjusted $p = 0.370$); the affected-ROI APE endpoint did not distinguish the two flow models.

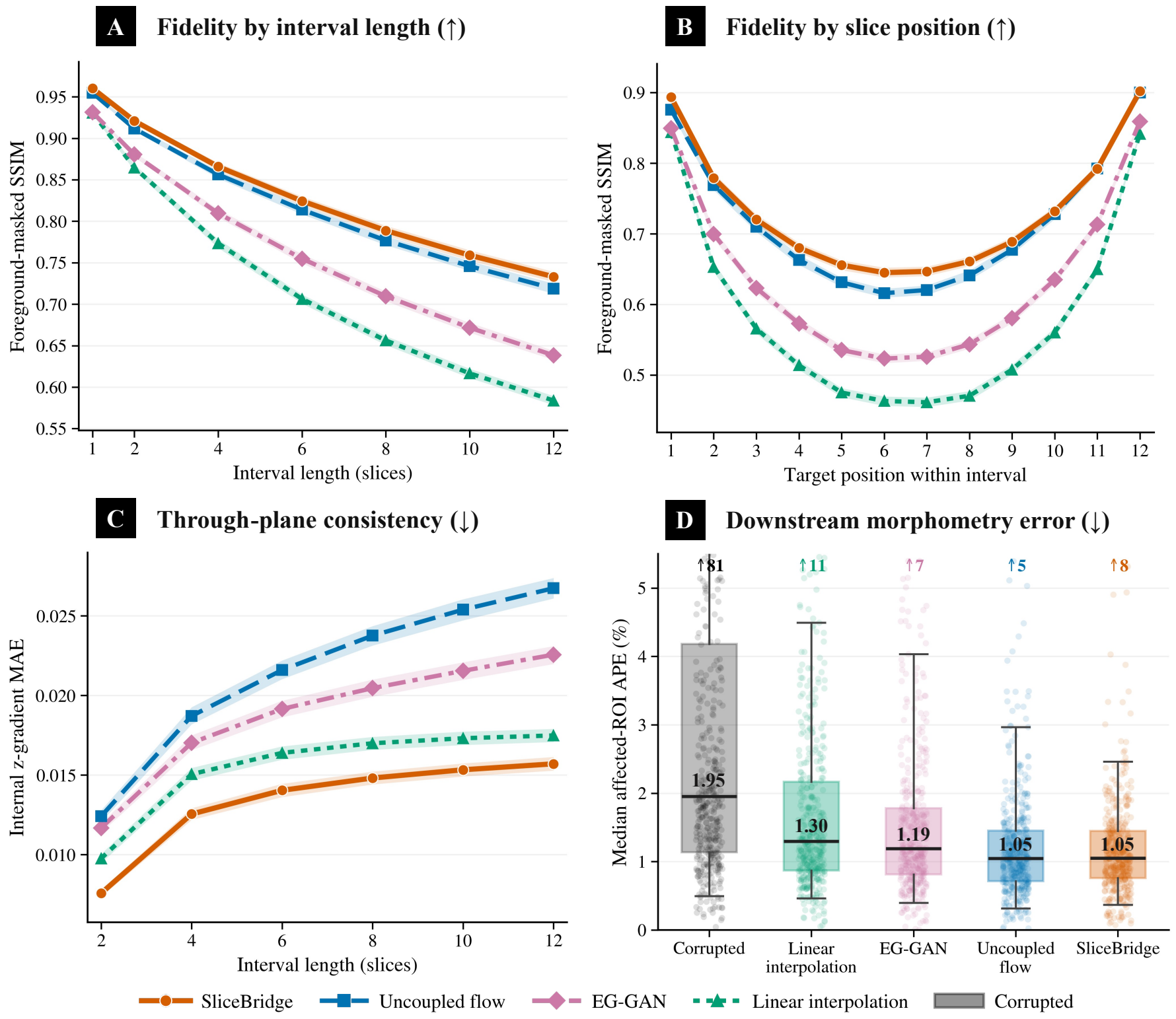

**Figure 3.** Quantitative evaluation on the external IXI cohort. (A) Foreground SSIM by interval length. (B) Position-wise SSIM for length-12 intervals, showing the greatest difficulty near the interval center. (C) Internal z-gradient MAE, where lower values indicate better within-interval through-plane consistency. (D) Median affected-ROI APE in controlled-corruption cases. Shaded bands in panels A-C denote 95% bootstrap confidence intervals. Arrows with numbers in panel D indicate the number of cases above the displayed y-axis limit.

### 4.4 Qualitative examples

Figure 4 shows a qualitative example from the external IXI cohort, in which the displayed target location remains unchanged, while the corrupted interval length was expanded from 1 to 6 and 12 slices. As the interval length increased, all methods exhibited increasing reconstruction distortion, but with different failure patterns. Uncoupled flow already showed ripple-like through-plane discontinuities in the coronal view at the medium interval, which became more pronounced at the longest interval. In this example, EG-GAN maintained relatively smooth coronal consistency but showed visibly lower axial reconstruction fidelity than the two flow models. SliceBridge better preserved the axial target appearance while reducing the through-plane discontinuities observed with Uncoupled flow.

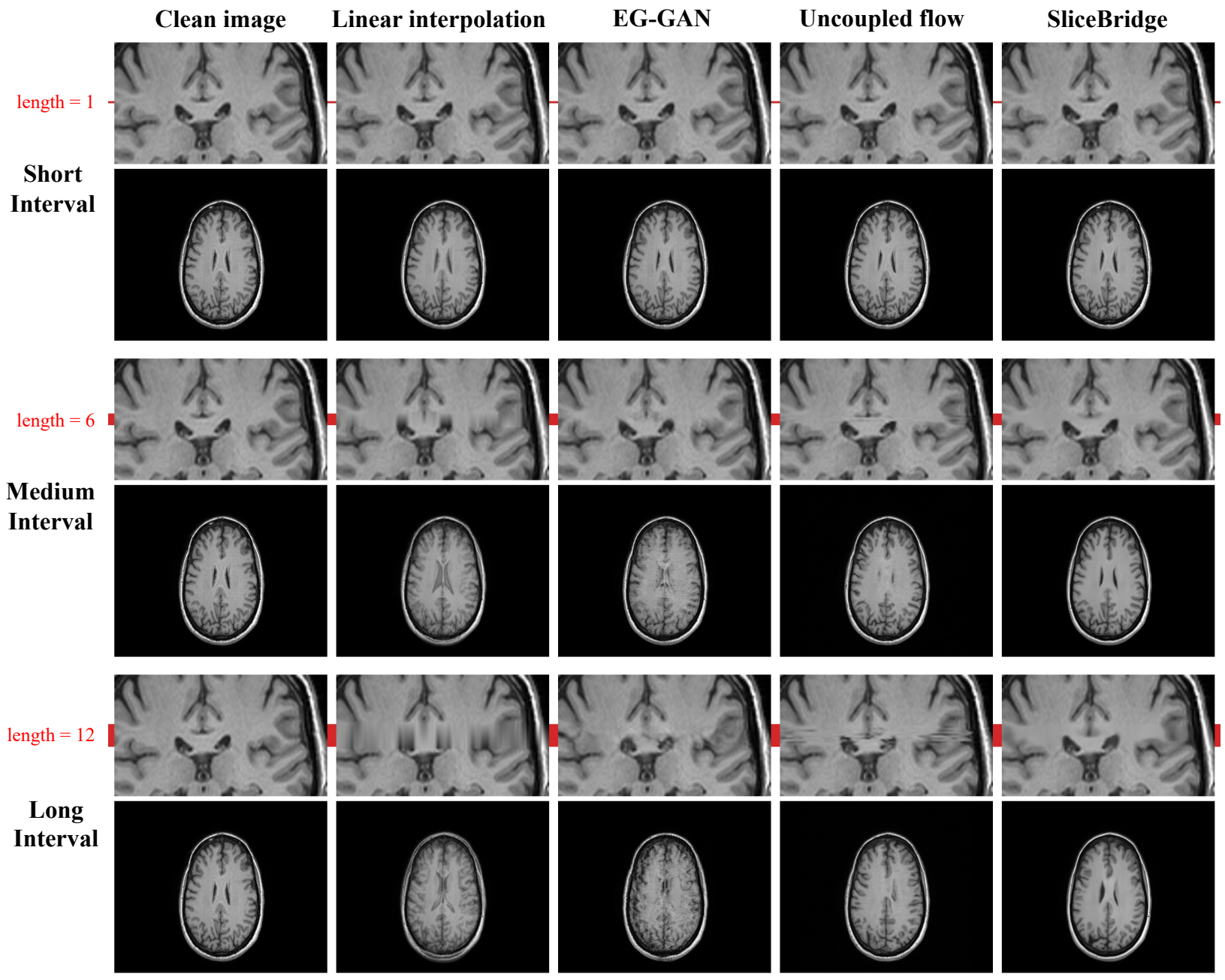

**Figure 4.** Qualitative comparison at a fixed target slice as the withheld interval expands from 1 to 6 and 12 slices. Each column shows the coronal reformat (top) and axial target (bottom). In this example, EG-GAN preserves smooth coronal consistency but loses axial fidelity, Uncoupled flow develops ripple-like through-plane discontinuities, and SliceBridge better balances axial appearance and through-plane consistency.

Figure 5 returns to the real localized slice corruption introduced in Figure 1 and compares the downstream SLANT-TICV segmentation before and after insertion of the SliceBridge repair candidate. After insertion of the SliceBridge repair candidate, the SLANT-TICV labels were more continuous across the corrupted interval and more consistent with those in neighboring slices. Because this real case has no verified clean reference, the comparison indicates improved consistency with downstream segmentation rather than recovery of the unknown original anatomy.

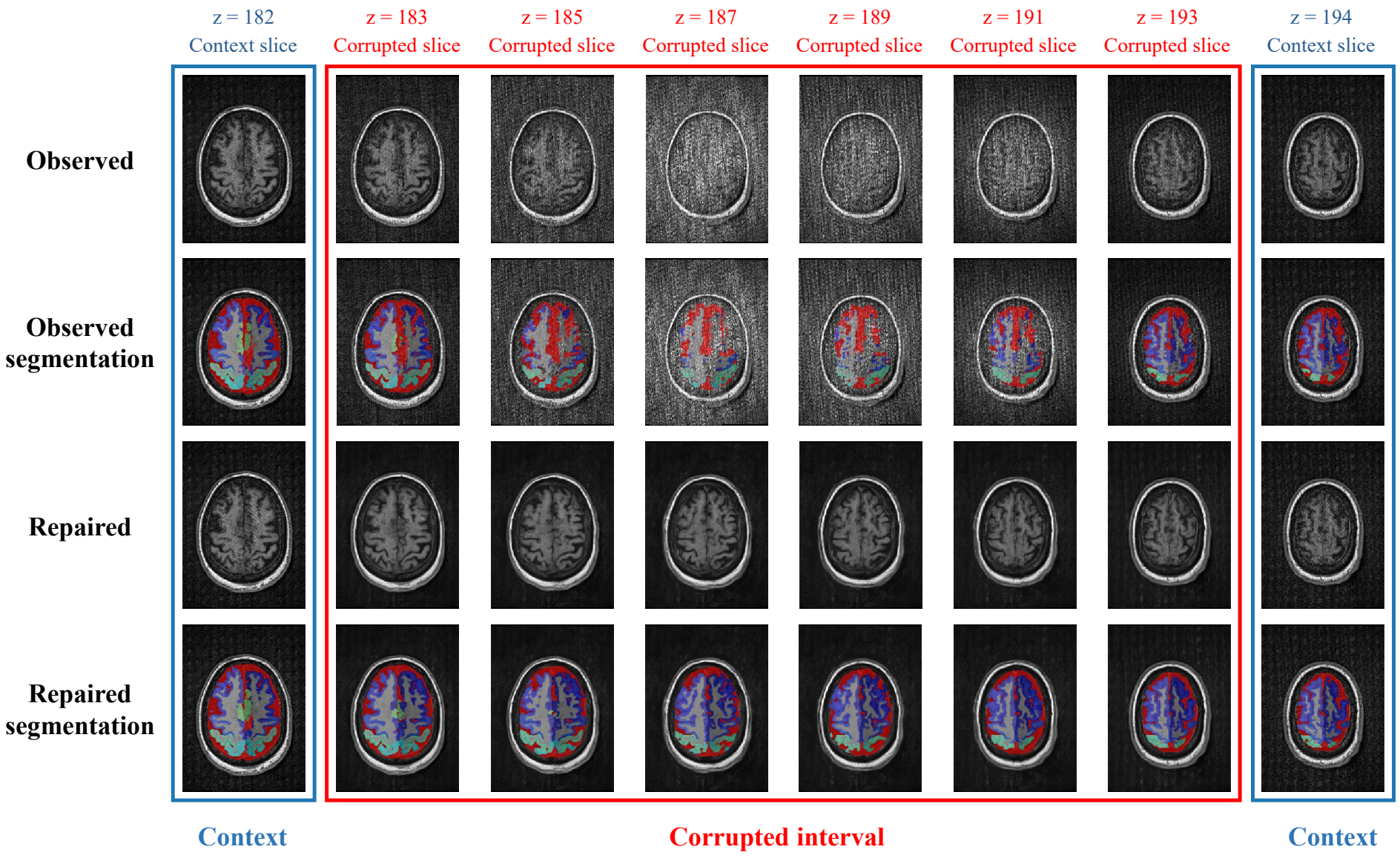

**Figure 5.** SliceBridge repair of the real corrupted slices introduced in Figure 1. Rows show the observed volume, its SLANT-TICV segmentation, the inserted repair candidate, and the resulting segmentation; blue borders denote context and red borders denote the repaired interval. The post-repair labels are more continuous across the interval.

## 5. DISCUSSION

Restoring a corrupted interval is not a problem of independently generating individual images. Each target must form a plausible complete axial image that is consistent with the surrounding context, while multiple reconstructed slices must also maintain a reasonable slice-to-slice progression. Uncoupled flow achieved high target-slice fidelity but still produced substantial discontinuities within the repaired interval. This finding shows that individually plausible slices do not necessarily form a coherent interval. The EG-GAN results further demonstrate that completing the repair in a coronal reformat does not automatically yield high axial fidelity or strong through-plane consistency. Repairing a complete interval therefore requires both reliable reconstruction of individual targets and joint coordination across multiple targets.

The morphometry experiment further delineated the respective roles of these two requirements. Both flow models reduced the morphometric error observed in corrupted volumes, but no significant difference was detected between them. Because the two models shared the same slice-wise rectified-flow formulation, while SliceBridge additionally incorporated interval-level coordination, these results suggest that the added effect of coordination was evident in through-plane consistency but was not reflected in this regional volume-based metric.

Several design choices define the scope of this proof-of-concept study and motivate direct extensions. SliceBridge assumes that unreliable intervals have already been identified; integrating automated detection remains future work. The current implementation reconstructs slices only in the axial plane, leaving coronal, sagittal, and multi-axis repair to future work. Requiring 16 intact context slices on both sides restricts the method to locations with complete bilateral context, which variable-length or one-sided conditioning could relax to reach the volume boundaries and more peripheral structures, including the cervical spinal cord[30]. Each interval was repaired with a single deterministic candidate, so repair uncertainty and output stability across random samples were not characterized. Finally, quantitative validation relied on clean-interval withholding and controlled corruptions. Future studies can evaluate manually confirmed slice corruptions to establish performance on real-world failures. Additional downstream measures sensitive to through-plane consistency may better capture the specific benefit of interval coordination.

Taken together, SliceBridge maintained high axial target fidelity while improving slice-to-slice consistency both within repaired intervals and across their boundaries, and reduced corruption-induced morphometric error. The work provides a basis for repository-level quality control systems that combine slice-level detection with context-matched repair candidates for human review.

## ACKNOWLEDGEMENTS

This work was supported by the Alzheimer’s Disease Sequencing Project Phenotype Harmonization Consortium (ADSP-PHC), funded by the National Institute on Aging under awards U24 AG074855, U01 AG068057, and R01 AG059716. This work was also supported by NIH awards R01 EB017230, K01 AG073584, and K01 EB032898. This work was conducted in part using the resources of the Advanced Computing Center for Research and Education at Vanderbilt University in Nashville, Tennessee.

Data collection and sharing for the Alzheimer's Disease Neuroimaging Initiative (ADNI) is funded by the National Institute on Aging (National Institutes of Health Grant U19 AG024904). The grantee organization is the Northern California Institute for Research and Education. In the past, ADNI has also received funding from the National Institute of Biomedical Imaging and Bioengineering, the Canadian Institutes of Health Research, and private sector contributions through the Foundation for the National Institutes of Health (FNIH) including generous contributions from the following: AbbVie, Alzheimer’s Association; Alzheimer’s Drug Discovery Foundation; Araclon Biotech; BioClinica, Inc.; Biogen; Bristol-Myers Squibb Company; CereSpir, Inc.; Cogstate; Eisai Inc.; Elan Pharmaceuticals, Inc.; Eli Lilly and Company; EuroImmun; F. Hoffmann-La Roche Ltd and its affiliated company Genentech, Inc.; Fujirebio; GE Healthcare; IXICO Ltd.; Janssen Alzheimer Immunotherapy Research & Development, LLC.; Johnson & Johnson Pharmaceutical Research &Development LLC.; Lumosity; Lundbeck; Merck & Co., Inc.; Meso Scale Diagnostics, LLC.; NeuroRx Research; Neurotrack Technologies; Novartis Pharmaceuticals Corporation; Pfizer Inc.; Piramal Imaging; Servier; Takeda Pharmaceutical Company; and Transition Therapeutics.

The NACC database is funded by NIA/NIH Grant U24 AG072122. NACC data are contributed by the NIA-funded ADRCs: P30 AG062429 (PI James Brewer, MD, PhD), P30 AG066468 (PI Oscar Lopez, MD), P30 AG062421 (PI Bradley Hyman, MD, PhD), P30 AG066509 (PI Thomas Grabowski, MD), P30 AG066514 (PI Mary Sano, PhD), P30 AG066530 (PI Helena Chui, MD), P30 AG066507 (PI Marilyn Albert, PhD), P30 AG066444 (PI John Morris, MD), P30 AG066518 (PI Jeffrey Kaye, MD), P30 AG066512 (PI Thomas Wisniewski, MD), P30 AG066462 (PI Scott Small, MD), P30 AG072979 (PI David Wolk, MD), P30 AG072972 (PI Charles DeCarli, MD), P30 AG072976 (PI Andrew Saykin, PsyD), P30 AG072975 (PI David Bennett, MD), P30 AG072978 (PI Ann McKee, MD), P30 AG072977 (PI Robert Vassar, PhD), P30 AG066519 (PI Frank LaFerla, PhD), P30 AG062677 (PI Ronald Petersen, MD, PhD), P30 AG079280 (PI Eric Reiman, MD), P30 AG062422 (PI Gil Rabinovici, MD), P30 AG066511 (PI Allan Levey, MD, PhD), P30 AG072946 (PI Linda Van Eldik, PhD), P30 AG062715 (PI Sanjay Asthana, MD, FRCP), P30 AG072973 (PI Russell Swerdlow, MD), P30 AG066506 (PI Todd Golde, MD, PhD), P30 AG066508 (PI Stephen Strittmatter, MD, PhD), P30 AG066515 (PI Victor Henderson, MD, MS), P30 AG072947 (PI Suzanne Craft, PhD), P30 AG072931 (PI Henry Paulson, MD, PhD), P30 AG066546 (PI Sudha Seshadri, MD), P20 AG068024 (PI Erik Roberson, MD, PhD), P20 AG068053 (PI Justin Miller, PhD), P20 AG068077 (PI Gary Rosenberg, MD), P20 AG068082 (PI Angela Jefferson, PhD), P30 AG072958 (PI Heather Whitson, MD), P30 AG072959 (PI James Leverenz, MD).

Data were provided in part by OASIS-3: Longitudinal Multimodal Neuroimaging: Principal Investigators: T. Benzinger, D. Marcus, J. Morris; NIH P30 AG066444, P50 AG00561, P30 NS09857781, P01 AG026276, P01 AG003991, R01 AG043434, UL1 TR000448, R01 EB009352. AV-45 doses were provided by Avid Radiopharmaceuticals, a wholly owned subsidiary of Eli Lilly.

The Amsterdam Open MRI Collection (AOMIC) is a collection of three datasets with multimodal (3T) MRI data including structural (T1-weighted), diffusion-weighted, and (resting-state and task-based) functional BOLD MRI data, as well as detailed demographics and psychometric variables from a large set of healthy participants. All raw data are publicly available from the OpenNeuro data sharing platform: ID1000: https://openneuro.org/datasets/ds003097, PIOP1: https://openneuro.org/datasets/ds002785, PIOP2: https://openneuro.org/datasets/ds002790/versions/2.0.0. We use version 1.2.1 for ID1000 and 2.0.0 for PIOP1 and PIOP2.

The IXI dataset is a publicly available brain MRI dataset.

During preparation of this manuscript, the authors used OpenAI ChatGPT to assist with language editing and stylistic refinement. All suggestions were reviewed and revised by the authors, who take full responsibility for the final manuscript.